\documentclass[10pt,twocolumn,letterpaper]{article}
\usepackage{wacv}
\usepackage{times} 
\usepackage{mathtools}
\usepackage{amssymb}
\usepackage{graphicx}
\usepackage{amsmath}
\usepackage{stmaryrd}
\usepackage{subcaption}
\usepackage{color}
\usepackage{booktabs}
\usepackage[ruled,vlined]{algorithm2e}

\usepackage[pagebackref=true,breaklinks=true,letterpaper=true,colorlinks,bookmarks=false]{hyperref}

\wacvfinalcopy 

\newcommand{\venue}[1]{{\small{#1}}}
\makeatletter
\newcommand{\fixed@sra}{$\vrule height 2\fontdimen22\textfont2 width 0pt\shortrightarrow$}
\newcommand{\shortarrow}[1]{%
  \mathrel{\text{\rotatebox[origin=c]{\numexpr#1*45}{\fixed@sra}}}
}
\makeatother

\def\wacvPaperID{355} 

\ifwacvfinal\fi
\begin{document}

\title{Adversarial Training of Variational Auto-encoders for \\ High Fidelity Image Generation}

\author{Salman H. Khan$^{\dagger }$, Munawar Hayat $^{\ddagger}$, Nick Barnes $^{\dagger }$ \\ 
 $^\dagger$Data61 - CSIRO and ANU, Australia, $^\ddagger$University of Canberra, Australia, \\
{\tt\footnotesize \{salman.khan,nick.barnes\}@data61.csiro.au, munawar.hayat@canberra.edu.au} 
}

\maketitle
\ifwacvfinal\thispagestyle{empty}\fi

\begin{abstract}
Variational auto-encoders (VAEs) provide an attractive solution to image generation problem. However, they tend to produce blurred and over-smoothed images due to their dependence on pixel-wise reconstruction loss. This paper introduces a new approach to alleviate this problem in the VAE based generative models. Our model simultaneously learns to match the data, reconstruction loss and the latent distributions of real and fake images to improve the quality of generated samples. To compute the loss  distributions, we introduce an auto-encoder based discriminator model which allows an adversarial learning procedure. The discriminator in our model also provides perceptual guidance to the VAE by matching the learned similarity metric of the real and fake samples in the latent space. To stabilize the overall training process, our model uses an error feedback approach to maintain the equilibrium between competing networks in the model. Our experiments show that the generated samples from our proposed model exhibit a diverse set of attributes and facial expressions and scale up to high-resolution images very well.
\end{abstract}

\section{Introduction}

Recent advances in deep learning have seen significant success in discriminative modeling for a wide range of classification tasks \cite{khan2017scene,khan2016discriminative,hayat2016spatial,hayat2016empoweringl,khan2017cost}. Generative models, however, still face many challenges in modeling complex data in the form of images and videos. Despite being fairly challenging, generative modeling of images is desirable in many applications. These include unsupervised and semi-supervised feature learning from large-scale visual data \cite{radford2015unsupervised}, understanding the representations learned by the discriminative models \cite{zeiler2010deconvolutional}, image completion \cite{yeh2016semantic}, denoising, super-resolution \cite{ledig2016photo} and prediction of future frames in a video \cite{vondrick2016generating}. Auto-encoder based models have traditionally been used for the generative modeling task. Variational Auto-Encoders (VAEs) are their improved variants which restrict the learned latent space representation to a prior probability distribution \cite{kingma2013auto}. The VAE based models approximate the data likelihood very well, however, their generated images are of low quality, do not retain fine details and have a limited diversity. 

Generative adversarial networks (GANs) provide a viable solution to the low quality output from VAEs \cite{goodfellow2014generative}. In theory, GANs can more accurately estimate the data distribution given an infinite amount of data and generate more realistic images. However in practice, GANs are difficult to optimize due to the lack of a closed-form loss function and can generate visually absurd outputs \cite{arjovsky2017towards}. Wasserstein metric \cite{arjovsky2017wasserstein} and energy based \cite{zhao2016energy} adversarial variants have been proposed to reduce the instability of GANs. A common limitation of all these approaches is the lack of control over the latent representations and therefore making it difficult to generate data with the desired attributes. 
To resolve this issue, \cite{larsen2015autoencoding} introduced an adversarial loss function to train VAEs. Though this helps stabilize the training and gives control over the latent distribution, the generated images are not sharp and crisp enough compared to their counterpart GAN based approaches. 

In this paper, we propose to match the data as well as the reconstruction loss and latent distributions for real and fake samples in the VAE during the training process. This allows us to recover high quality image samples while having full control over the latent representations. Leveraging on the learned latent space for both real and fake images, we introduce priors to ensure the generation of perceptually plausible outputs. The complete model is trained using an adversarial loss function which allows us to learn a rich similarity measure for images. This leads to a highly flexible and robust VAE architecture, that combines the benefits of both variational and adversarial generative models. Furthermore, drawing insight from the recent boundary equilibrium GAN \cite{berthelot2017began}, we design a controller to balance the game between the generator and the discriminator. This results in a smooth optimization and allows us to avoid the commonly employed heuristics for stable training.

In summary, this paper makes the following contributions:
1) We propose a new VAE+GAN model, capable of generating high fidelity images. The proposed model encodes images in a compact and meaningful latent space where different arithmetic operations can be performed and reflected back in the image domain.
2) Our proposed model incorporates a learned similarity metric to avoid unrealistic outputs and generates globally coherent images which are perceptually more appealing.
3)  To stabilize the model training, we propose to consider the past and future trends of the error signal obtained by comparing the discriminator and generator losses.

Next, we outline the related work followed by our proposed model.

\section{Related Work}
Generative image modeling using dictionary learning techniques have been extensively studied in the literature with applications to texture synthesis \cite{efros1999texture}, in-painting \cite{hays2007scene} and image super-resolution \cite{freeman2002example}. In comparison, generating natural images did not see much success until recently with advances in deep learning. Restricted Boltzmann machines \cite{hinton1986learning} and deep auto-encoder models \cite{hinton2006reducing} are amongst the earliest neural networks based methods for unsupervised feature learning and image generation. These models first encode images into a latent space and then decode them back into image space by minimizing pixel-wise reconstruction errors. Kingma and Welling \cite{kingma2013auto} proposed a variational inference based encoding-decoding approach which enforces a prior on the learnt latent space. Their proposed method achieved promising results but the generated images were often blurry.

Auto-encoder based models are trained to minimize pixel-level reconstruction errors. These models do not consider holistic contents in an image as interpreted by human visual perception. For example, a small scale rotation would yield large pixel-wise reconstruction error, but would be barely noticeable to human visual perception. Generative Adversarial Networks  \cite{goodfellow2014generative} can learn a better similarity metric for images, and have received significant research attention since their introduction. A GAN comprises two network modules: a generator which maps a sample from a random uniform distribution into an image space, and a discriminator which predicts an image to be real (from the database) or fake (from the generator). Both modules are trained with conflicting objectives based upon the principles of game theory. Compared with the previous approaches which were mainly based upon minimizing pixel-wise reconstruction errors, discriminators in GANs learn a rich similarity metric to discriminate images from non-images. GANs can therefore generate promising and sharper images \cite{radford2015unsupervised}. GANs are however unstable to train and the generated images are prone to being noisy and incomprehensible. Since their release, many efforts have been made to improve GANs. Radford \etal \cite{radford2015unsupervised} were the first to incorporate a convolutional architecture in GANs, which resulted in improved quality of the generated images. Incorporating side information (class labels) into the discriminator module of GANs has also been shown to improve the quality of generated images \cite{odena2016conditional}.

GANs still face many challenges: they are difficult to train and require careful hyper-parameter selection. While training, they can easily suffer from modal collapse \cite{dumoulin2016adversarially}, a failure mode in which they generate only a single image. Further, it is quite challenging to best balance the convergence of the discriminator and the generator, since the discriminator often easily wins at the beginning of training. Salimans \etal \cite{salimans2016improved} proposed architectural improvements that stabilize the training of GANs. In order to strike a balance between the generator and discriminator, Boundary Equilibrium GANs (BEGAN) \cite{berthelot2017began} recently introduced an equilibrium mechanism. BEGANs  gradually change the emphasis of generator and discriminator loss terms in gradient descent as the training progresses. Zhao \etal proposed Energy based GANs (EBGANs) \cite{zhao2016energy} which model the discriminator as an energy function, and implement it as an auto-encoder. The discriminator energy function can be viewed as a trainable loss function for the generator. EBGANs are more stable and less prone to hyper-parameter selection. EBGANs and earlier GAN versions however lack a mechanism to estimate convergence of the trained model. More recently, Wasserstein GANs (WGANs) \cite{arjovsky2017wasserstein} introduced a loss that shows a correlation between discriminator loss and perceptual quality of the generated images. Wasserstein loss can therefore act as an indicator for model convergence.

Our work is a continuation of effort to devise stable deep generative models which produce diverse and improved quality images with high resolution. To this end, different from previous works, our proposed model learns to simultaneously match the data distributions, loss distributions and latent distributions of real and fake data during the training stage. We show that such a hierarchical but flexible supervision in the model helps generate images with better quality and high resolution. The closest to our approach is the BEGAN \cite{berthelot2017began} with notable differences including the maximization of a lower bound (Sec.~\ref{Minimizing Loss Distributions}), perceptual guidance in the latent space (Sec.~\ref{Perceptual Guidance}),  the combination of data and loss distribution matching in the training objective (Sec.~\ref{Adverserial Training}) and incorporation of VAEs in the generator module to learn a latent space where an image of desired style can deterministically be generated using vector arithmetics (Sec.~\ref{Exploring the Latent Space}).

\begin{figure*}[th]
\centering
   \begin{minipage}[c]{0.48\textwidth}
    \caption{\textbf{Model Overview:} The \textcolor{red}{red} dotted lines represent the loss functions, the downward diagonal arrows ($\protect\shortarrow{7}$) represent the decoding operation and the upwards diagonal arrows  ($\protect\shortarrow{1}$) represent the encoding operation. The proposed model first encodes the input image ($\mathbf{x}$) to a parameterized latent representation ($\mathbf{z}_v$) and reconstructs it back using the generator ($\mathbf{x}_g$ and $\mathbf{x}_v$ corresponding to latent representations $\mathbf{z}_g$ and $\mathbf{z}_v$ respectively). The discriminator also consists of an auto-encoder which first encodes the inputs to a latent representation ($\mathbf{z}_d$, $\mathbf{z}'_g$ and $\mathbf{z}'_v$ corresponding to inputs $\mathbf{x}$, $\mathbf{x}_g$ and $\mathbf{x}_v$  respectively), and then reconstructs them back ($\mathbf{x}'_d$, $\mathbf{x}'_g$ and $\mathbf{x}'_v$). The generator and discriminator are trained using the back-propagated error signal from the losses computed to match the actual data, the latent representation and the loss distributions for the real and fake samples.} \label{fig:overview}
  \end{minipage}
  \;
  \begin{minipage}[c]{0.5\textwidth}
    \includegraphics[clip=true, width = 1\textwidth]{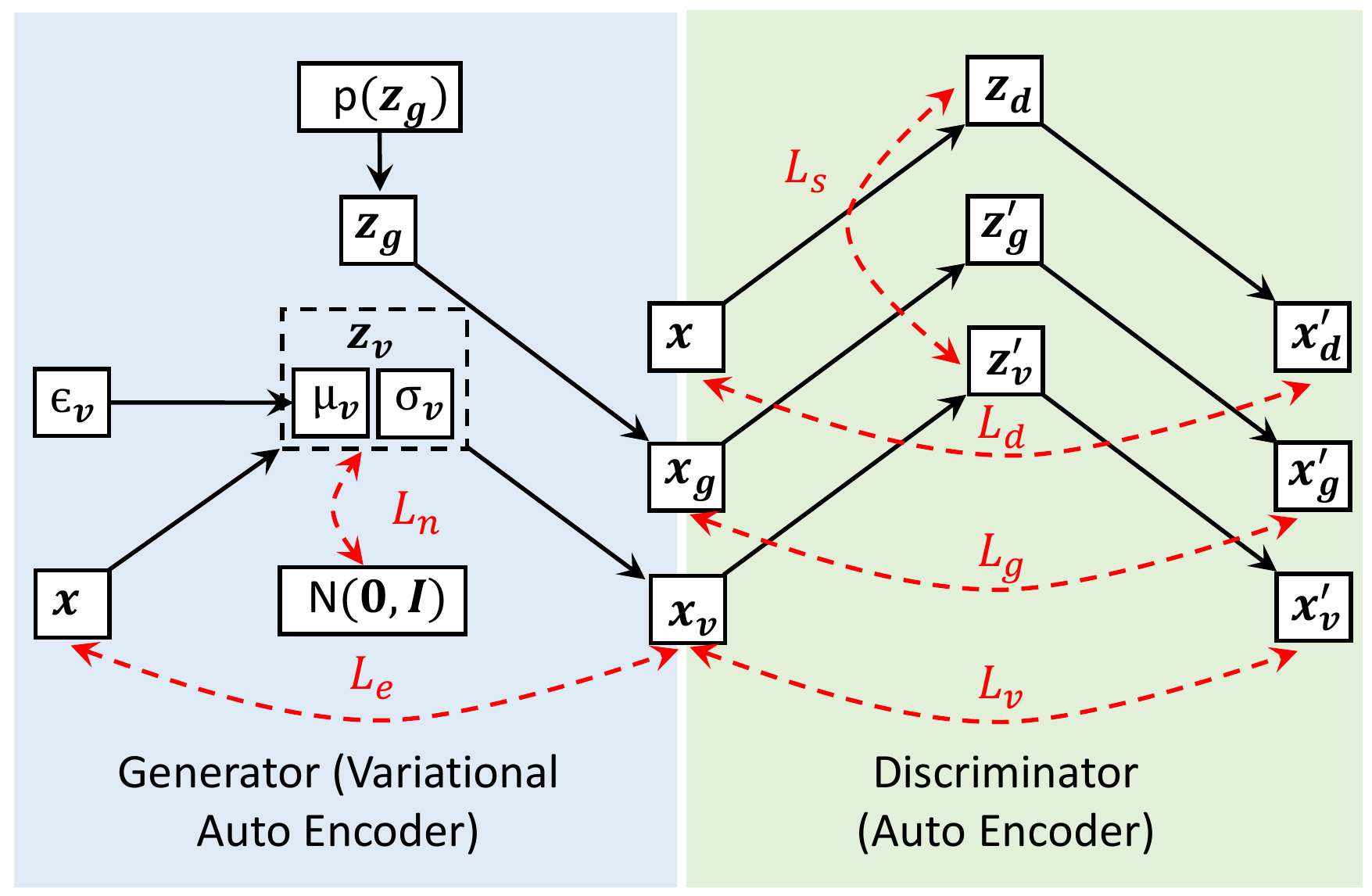}
  \end{minipage}\hfill
\end{figure*}

\section{Proposed Model}
Given a set of data samples $\mathcal{X} = \{\mathbf{x}_i: i \in [1,n]\}$ from an unknown data distribution, we want to learn a generative model with parameters $\theta$ that maximizes the likelihood:
\begin{align}
\theta^* = \underset{\theta \in \Theta}{\max} \sum_{i=1}^{n} \log \ell(\mathbf{x}_i; \theta),
\end{align}
where $\ell(\cdot)$ denotes the density of each observation $\mathbf{x}_i$. After training, the  model can then be used to obtain new samples from the learned distribution.

The proposed model consists of a pair of auto-encoders, as shown in Fig.~\ref{fig:overview}. In contrast to a traditional pixel-wise loss minimization over the data samples, the proposed approach minimizes the loss at the data level, in terms of reconstruction loss and the latent distributions of real and fake images. An input data sample is first encoded and decoded using the CNN blocks which together form a variational auto-encoder whose latent distribution is parameterized as a normal distribution. Afterwards, the reconstructed images are passed through another auto-encoder and the reconstruction error is minimized for both the original and fake images. The overall network is trained using an adversarial loss function which incorporates a learned perceptual similarity metric to obtain high-quality synthetic images. We begin with an overview of the variational auto-encoder, followed by the proposed improvements to enhance the visual quality of generated images.

\subsection{Variational Auto-encoder}
Auto-encoders prove a very powerful tool to model the relationship between data samples and their latent representations. A variational auto-encoder is similar to the regular auto-encoder in the sense that it first transforms the given data to a low-dimensional latent representation and then projects it back to the original data space. Given an input data sample $\mathbf{x}$, the encoding operation can be represented as:
$\mathcal{F}_{vae-e}(\mathbf{x}; \theta_e) = \mathbf{z}_v \sim q_{\theta}(\mathbf{z}_v|\mathbf{x})$, where $\mathbf{z}_v$ denotes the latent representation. The encoder function $\mathcal{F}_{vae-e}(\mathbf{x}; \theta_e)$  is implemented as a CNN with parameters $\theta_e$. Similarly, the decoding operation is implemented as another CNN and can be represented as:  $\mathcal{F}_{vae-d}(\mathbf{z}_v; \theta_d) = \mathbf{x}_v \sim p_{\theta}(\mathbf{x}|\mathbf{z}_v)$.  To be able to reconstruct the original data sample, the following loss function is minimized:
\begin{align}\label{eq:data_loss}
\mathcal{L}_e = \frac{1}{|\mathbf{x}|}\parallel \mathbf{x} - \mathcal{F}_{vae-d}(\mathcal{F}_{vae-e}(\mathbf{x} ; \theta_e); \theta_d) \parallel_1
\end{align}
Here, $\parallel \cdot \parallel_1$ denotes the $\ell_1$ norm and $|\cdot|$ denoes the latent space cardinality.
The main distinguishing factor of VAE in comparison to a vanilla auto-encoder is the constraint on the encoder to match the low-dimensional latent representation to a prior distribution. The regularization on the encoded latent representation means that the $\mathbf{z}_v$ is constrained to follow a unit Gaussian distribution i.e., $\mathbf{z}_v \sim N({0},{I})$. This is achieved by minimizing the Kullback-Leibler ($\mathrm{KL}$) divergence between the two distributions as follows:
\begin{align}\label{eq:l_n}
\mathcal{L}_n &= \mathrm{KL}( N({\mu}_v,\sigma_v) || N({0},{I}))
\end{align}

Since an end-to-end training is not possible with the intermediate stochastic step which involves sampling from $N(\mu_v, \sigma_v)$, the re-parametrization trick proposed by \cite{kingma2013auto} is used to enable the error back-propagation by treating the stochastic sampling as an input to the network. As a result, the latent variable is defined as:
\begin{align}
\mathbf{z}_v = \mu_v + \epsilon_v * \sigma_v, \quad \epsilon_v \sim N({0},I), 
\end{align}
where $\mu_v$ and $\sigma_v$ denote the mean and variance while $\epsilon_v$ is randomly sampled from a unit Gaussian distribution.
After the training process,  the decoder can be used independently to generate new data samples $\mathbf{x}_g$ by feeding randomly generated samples from the distribution: $\mathbf{z}_g \sim p(\mathbf{z}_g) = N({0},{I})$. The main problem with the VAE based models is that the per-pixel loss function $\mathcal{L}_e$ (Eq.~\ref{eq:data_loss}) defined in terms of mean error results in over-smoothed and blurry images. To overcome this problem, we propose to match the loss and latent distributions of real and fake images in addition to the minimization of commonly used mean per-pixel error measures. In the following, we first describe the loss distribution matching process and then outline the matching of a perceptual similarity metric using the latent distributions.

\subsection{Minimizing Loss Distributions}\label{Minimizing Loss Distributions}
The VAE set-up described above outputs a reconstructed data sample. The decoding stage in the VAE is similar to the generator function in a regular Generative Adversarial Network (GAN). The generator in a vanilla GAN is followed by a discriminator which distinguishes between the real and fake (generated) data samples. In essence, the distribution of fake and real data samples is matched by playing a game between the generator and the discriminator until the Nash equilibrium is reached. Inspired  by Berthelot \etal \cite{berthelot2017began}, in addition to directly matching the data distributions, we also minimize the approximate Wasserstein distance  between the reconstruction loss distributions of real and fake data samples. For sufficiently large number of pixels, the distributions will be approximately normal. The generator will be trained to minimize the reconstruction loss of the fake samples, thus trying to produce real looking samples so that the discriminator assigns them a lower energy. On the other hand, the discriminator will be trained to minimize the reconstruction error for real samples, but maximize the error for the fake samples coming from the generator.  We can represent the generator and discriminator loss functions as:
\begin{align}\label{eq:loss_1}
\mathcal{L}_{dis} &= \mathcal{L}_d - (\mathcal{L}_g + \alpha\mathcal{L}_v)\\
\mathcal{L}_{gen} & = \mathcal{L}_g + \alpha\mathcal{L}_v
\end{align}
where the individual loss terms are defined as: 
\begin{align}\label{eq:l_gvd}
\mathcal{L}_d = \frac{1}{|\mathbf{x}|}\parallel \mathbf{x} - \mathbf{x}'_d \parallel_1, & \quad
\mathcal{L}_g = \frac{1}{|\mathbf{x}_g|}\parallel \mathbf{x} - \mathbf{x}'_g \parallel_1, \notag\\
\quad \mathcal{L}_v = & \frac{1}{|\mathbf{x}_v|}\parallel \mathbf{x}_v - \mathbf{x}'_v \parallel_1.
\end{align}
Here $\alpha$ is a weighting parameter which decides the emphasis on the reconstruction loss of recovered training samples from the VAE and $\mathbf{x}'_d$, $\mathbf{x}'_g$ and $\mathbf{x}'_v$ denote the samples reconstructed by the discriminator auto-encoder corresponding to inputs $\mathbf{x}$, $\mathbf{x}_g$ and $\mathbf{x}_v$ respectively:
\begin{align}
\mathbf{x}'_{v,g,d} = \mathcal{F}_{ae-d}(\mathcal{F}_{ae-e}(\mathbf{x}_{v,g,-}; \theta_{e'}); \theta_{d'})
\end{align}
such that $\theta_{e'}, \theta_{d'}$ represent the parameters of discriminator encoder and decoder respectively. 
The above defined model can be understood as an improved version of the energy-based generative network \cite{zhao2016energy}, where the reconstruction error represents the energy assigned by the discriminator, with low energies being assigned to the samples close to the real data manifold.

We use a simple auto-encoder to estimate the loss distributions of real and fake data. This greatly stabilizes the generative training and avoids high sensitivity to hyper-parameters and undesirable learning modes such as the model collapse. Note that until now, we are matching the reconstructed output from the generator and the loss distributions of real and fake images.  This can lead to the generation of simple images which are easy to reconstruct by the discriminator. It is also important to encourage the generator to produce more complex, diverse and realistic samples. For this purpose, we propose to learn a perceptual metric which forces the generator to create more realistic and diverse images.  

\subsection{Perceptual Guidance} \label{Perceptual Guidance} 
In order to enhance the visual quality of generated images, we propose to add perceptual guidance in the latent space of the discriminator while training the output energy function. This acts as a regularizer during model training and enforces similarity between the real and generated samples using a learned metric. Assuming that the learned latent representation is a compact and perceptually faithful encoding of the real and generated samples, we enforce a loss term in the encoder and generator modules which measures the similarity as an $\ell_1$ norm of the difference of the latent representations in the discriminator corresponding to the real and fake images, denoted by $\mathbf{z}_d$ and $\mathbf{z}'_v$ respectively: 
\begin{align}\label{eq:l_s}
\mathcal{L}_s = \frac{1}{|\mathbf{z}_d|}\parallel\mathbf{z}_d - \mathbf{z}'_v\parallel_1.
\end{align}
This loss essentially encourages the generator to output images which are close to the data manifold of real images. This is achieved by measuring the similarity between fake and real images in a more abstract sense which aims to roughly match the style of the two image types. Note that we also include a content loss in the the generator training objective to directly minimize the reconstruction error in the VAE model (Eq.~\ref{eq:data_loss}). The combination of the loss computed in the latent space of the discriminator and the content loss in the VAE model complement each other and result in an optimal training of the model. We describe our adversarial training approach to train the overall model in the next section.

\subsection{Adversarial Training}\label{Adverserial Training}
The discriminator in Eq.~\ref{eq:loss_1} contains two objectives i.e. to accurately reconstruct real images (formulated as the loss $\mathcal{L}_d$) and to distinguish between the real and fake samples (by maximizing the distance between $\mathcal{L}_d$ and $\mathcal{L}_g$). It is necessary to keep the right balance between these two objectives for an optimal model training. The boundary equilibrium technique in \cite{berthelot2017began} used a  proportional controller to maintain this balance using a feedback signal. This feedback signal is defined in terms of the error between the weighted reconstruction loss of real and fake data samples. It maintains the equilibrium such that the expected values of both losses are balanced by a constant factor, termed as the diversity factor:
\begin{align}\label{eq:10}
\eta = \frac{\mathbb{E}[\mathcal{L}_g] + \alpha\mathbb{E}[\mathcal{L}_v]}{\mathbb{E}[\mathcal{L}_d]}
\end{align}

\begin{figure*}[htp]
	\centering
	\includegraphics[width=\textwidth]{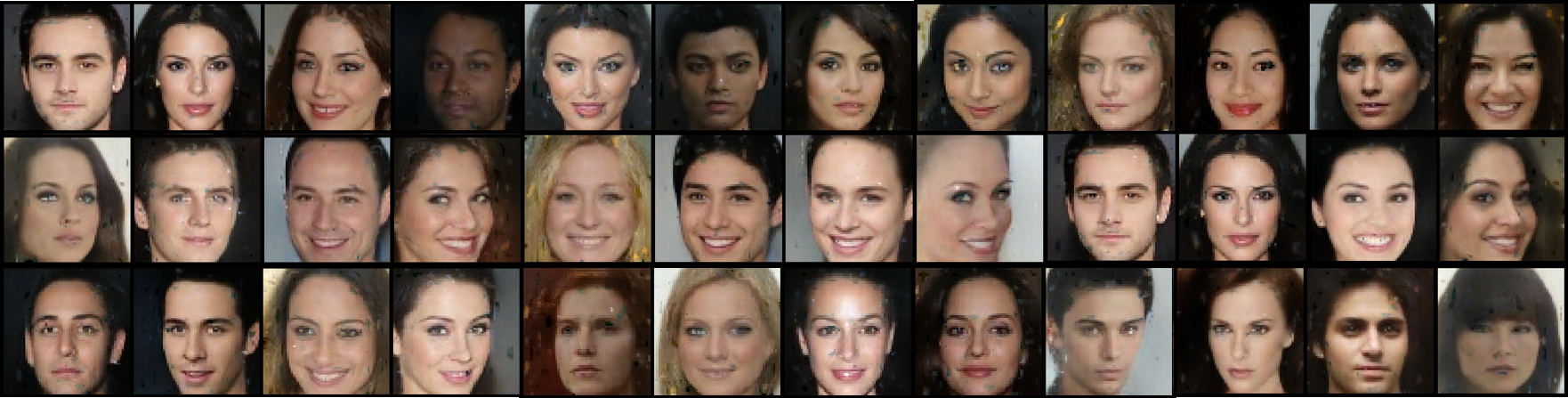}
	\caption{Generated Images ($64 \times 64$)}
	\label{fig:results_64}
\end{figure*}

\begin{figure}[t] 
	\centering
	\includegraphics[width=0.99\columnwidth]{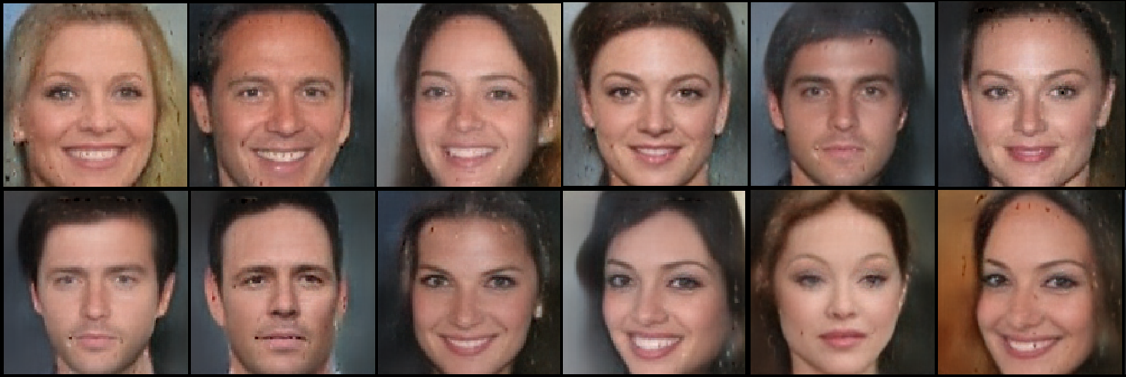}
	\caption{Generated Images ($128 \times 128$)}
	\label{fig:results_128}
\end{figure}

In practice, the proportional control mechanism slowly attains equilibrium when the proportional gain factor is too low. Setting the gain to a higher value leads to unstable learning procedure. Furthermore, since the update is driven by the error signal calculated using the feedback, an exact equilibrium is never attained during the training rather a steady state error is maintained. To avoid these problems, we introduce the accumulated error signal to accelerate the attainment of equilibrium and to avoid the steady state error. We noticed that the accumulated error from the feedback signal can easily cause  oscillations around the reference signal. This causes instability during the training and results in suboptimal generated samples. To overcome this problem, we also introduce a differential term while updating the  equilibrium parameter `$k_t$'. It dampens the oscillations around the reference signal without affecting the signal matching time. In order to stabilize the system, we obtain the derivative term using multiple error samples and use a small differential gain in the update equation as follows:
\begin{align}
e_t =& \eta \mathcal{L}_d - \mathcal{L}_g + \alpha \mathcal{L}_v \qquad \text{at iteration } t \label{eq:11}\\
k_{t} =& k_{t-1} + \lambda_1 e_t  + \lambda_2 (e_t - e_{t-1}) + \lambda_{3} (e_t + e_{t-2} - 2e_{t-1}) \label{eq:12}
\end{align}
where, $e_t$ denote the error term, $\lambda_{1-3}$ denote the gain parameters for the integral, proportional and differential components. 
The overall loss function can therefore be expressed as:
\begin{align}
\mathcal{L}_{dis} & = \mathcal{L}_d - k_{t-1} (\mathcal{L}_g + \alpha\mathcal{L}_v) \label{eq:l_dis}\\
\mathcal{L}_{gen} & = \mathcal{L}_g + \alpha\mathcal{L}_v + \beta\mathcal{L}_s + \gamma\mathcal{L}_e \label{eq:l_gen}\\
\mathcal{L}_{enc} & = \mathcal{L}_n  + \beta\mathcal{L}_s + \gamma\mathcal{L}_e \label{eq:l_enc}
\end{align}
where, $\alpha, \beta, \gamma$ denote the weights which put different level of emphasis on the respective loss functions. 
Note that our generator and discriminator loss functions incorporate the reconstruction loss $\mathcal{L}_v$ computed on the real data samples reconstructed by the VAE. This forces the generator to produce high quality samples lying close to the real data manifold. Similarly, the generator model is trained on the data loss ($\mathcal{L}_e$), the reconstruction loss and the latent space similarity loss ($\mathcal{L}_s$). The convergence rate of the model can be measured by analyzing the error measure given by $\mathcal{L}_d + |\eta \mathcal{L}_d - \mathcal{L}_g - \alpha\mathcal{L}_v|$. The overall training is illustrated in Algorithm~\ref{Algorithm}.

\begin{algorithm}[t]
\nl {\bf Initialization:} $\theta_e, \theta_d, \theta_{e'}, \theta_{d'} \leftarrow $ randomly initialize VAE and AE encoder, decoder respectively. \\
// Perform a total of $T$ training iterations \\
\For{$t = 1:T$ }{
 	{\bf VAE Training}\\
	\nl  $\mathbf{X} \leftarrow$ sample a random batch from $\mathcal{X}$ \\
    \nl $\mathbf{Z}_v \leftarrow \mathcal{F}_{vae-e}(\mathbf{X}; \theta_e)$\\
    \nl $\mathbf{Z}_g \sim N(\mathbf{0},\mathbf{I}) $ \\
    \nl $\mathbf{X}_{v,g} \leftarrow \mathcal{F}_{vae-d}(\mathbf{Z}_{v,g}; \theta_d)$ \\
    \nl $\mathbf{Z}'_{v,g,d} \leftarrow \mathcal{F}_{ae-e}(\mathbf{X}_{v,g,-}; \theta_{e'})$ \\
    \nl Calculate $\mathcal{L}_e$, $\mathcal{L}_n$ and $\mathcal{L}_s$ using Eqs.~\ref{eq:data_loss}, \ref{eq:l_n} and \ref{eq:l_s}\\
    \nl $\theta_e \leftarrow \nabla_{\theta_e} \mathcal{L}_{enc}$ (Eq.~\ref{eq:l_enc})\\
    {\bf Adversarial Training}\\
    
    \nl $\mathbf{X}'_{v,g,d} \leftarrow \mathcal{F}_{ae-d}(\mathbf{Z}'_{v,g,d}; \theta_{d'})$ \\
    \nl Calculate $\mathcal{L}_g, \mathcal{L}_d$ and $ \mathcal{L}_v$  using Eq.~\ref{eq:l_gvd}\\
    \nl $\theta_d \leftarrow \nabla_{\theta_d} \mathcal{L}_{gen}$ (Eq.~\ref{eq:l_gen})\\
    \nl Calculate $\eta, e_t$ and $k_t$ using Eqs.~\ref{eq:10}, \ref{eq:11} and \ref{eq:12}\\
    \nl $\theta_{e'}, \theta_{d'} \leftarrow \nabla_{\theta_{e'}, \theta_{d'}} \mathcal{L}_{dis}$ (Eq.~\ref{eq:l_dis})\\
    }
{\bf Return:} Updated parameters $\theta_e, \theta_d, \theta_{e'}, \theta_{d'}$
    
    \caption{{\bf Learning procedure for proposed model} \label{Algorithm}}

\end{algorithm}

\section{Implementation Details}
Both the generator and discriminator in our proposed model consist of an encoder and a decoder. For simplicity, we keep the backbone architecture of both the generator and the discriminator identical, except that the generator models a VAE. With in the generator and the discriminator, the architecture of encoder and decoder are also equivalent to each other in terms of the number of layers and therefore the parameters. 
We keep the design of encoder and decoder consistent with the discriminator module of \cite{berthelot2017began}
This design choice was made due to two reasons: \textbf{(1)} The encoder and decoder architectures are fairly simple compared to their counterpart models used in GANs. These consist of three pairs of convolution layers, each with an exponential linear unit (ELU) non-linearity and interleaved with sub-sampling and up-sampling layers for the case of encoder and decoder respectively. Furthermore, the architecture does not use dropout, batch-normalization and convolution transpose layers as for the case of other competing model architectures. \textbf{(2)} It makes it easy to compare our model with the latest state of the art BEGAN model \cite{berthelot2017began}, which achieves good results in terms of visual quality of the generated images. Note that different from \cite{berthelot2017began}, our generator module is based upon a VAE and both the generator and the discriminator takes into account data, loss and latent distributions of real and fake data.

The latent vector $\mathbf{z} \in \mathbb{R}^N$ is sampled from $[0,1]$  following a normal distribution. During training, the complete model shown in Fig.~\ref{fig:overview} is used. At test time, the encoder and the discriminator are discarded and random samples from $p(\mathbf{z}_g)$ are feed forwarded through generator to obtain new data samples. We use the gain parameters settings to be $10^{-3}, 10^{-5}$ and $10^{-5}$ respectively  in our experiments. The parameters $\alpha, \beta, \gamma$ and $\eta$ are set to $0.3, 0.1, 0.1$ and $0.5$ respectively. The model was trained for $300$k iterations using the Adam optimizer initialized with a small learning rate of $5\times 10^{-5}$.

\section{Experiments}
\subsection{Dataset}
We use the CelebFaces Attribute Dataset (CelebA) \cite{liu2015faceattributes} in our experiments. CelebA is a large scale dataset with more than $202$k face images of over $10$k celebrities. The dataset contains a wide range of appearances, head poses and backgrounds. The dataset has been annotated with 40 attributes, however, these are not used in our unsupervised training of the proposed model. We use the aligned and cropped version of images in the dataset where the face appears roughly at the center of an image.

\begin{table}[htp]
\centering
\begin{tabular}{c c}
\toprule
Approach & Inception Score  \\
\midrule
DCGAN \venue{(ICLR'16)} \cite{radford2015unsupervised} & 4.89\\
Improved GAN \venue{(NIPS'16)} \cite{salimans2016improved} & 4.36\\
ALI \venue{(ICLR'17)} \cite{dumoulin2016adversarially} & 5.34\\
MIX + WGAN (ICML'17) \cite{arora2017generalization} & 4.04\\
PixelCN++ \venue{(ICLR'17)} \cite{salimans2017pixelcnn++} & 5.51 \\
AS-VAE-g \venue{(NIPS'17)} \cite{pu2017adversarial} & \textcolor{red}{\bf 6.89} \\
BEGAN \venue{(Arxiv'17)} \cite{berthelot2017began} & 5.62\\
Ours (BEGAN* + LSM) & 6.12\\
Ours (VAE + BEGAN* + LSM) & \textcolor{blue}{\bf 6.80} \\
\midrule
Improved GAN (semi-supervised)  \cite{salimans2016improved} & 8.09 \\
Real Images & 11.24 \\
\bottomrule
\end{tabular}
\caption{Quantitative comparison on the CIFAR-10 dataset in terms of Inception score. The best and the second best performances are shown in \textcolor{red}{red} and \textcolor{blue}{blue} respectively. BEGAN* denotes \cite{berthelot2017began} with the modified equilibrium  approach and LSM stands for the Learned Similarity Metric. All the reported performances are for unsupervised cases except the bottom two, which use label information or real images, respectively. }
\label{tab:quan_comp}
\end{table}

\begin{figure*}[htp]
	\centering
	\begin{subfigure}[t]{0.5\textwidth}
		\centering
		\includegraphics[width=0.98\textwidth]{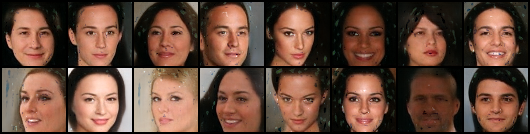}
		\caption{Fake Images from the Generator}
	\end{subfigure}%
	~ 
	\begin{subfigure}[t]{0.5\textwidth}
		\centering
		\includegraphics[width=0.98\textwidth]{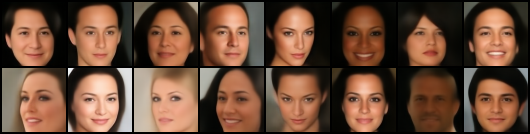}
		\caption{Reconstructed fake images (discriminator output)}
	\end{subfigure}
	~
	\begin{subfigure}[t]{0.5\textwidth}
		\centering
		\includegraphics[width=0.98\textwidth]{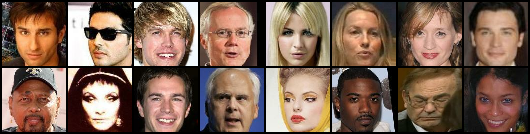}
		\caption{Real Images}
	\end{subfigure}%
	~ 
	\begin{subfigure}[t]{0.5\textwidth}
		\centering
		\includegraphics[width=0.98\textwidth]{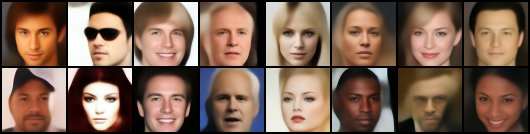}
		\caption{Reconstructed real images (discriminator output)}
	\end{subfigure}
	\caption{Real and fake images and their corresponding reconstructions by the discriminator.}
	\label{fig:reconstructions}
\end{figure*}

\begin{figure*}[htp]
\centering
   \begin{minipage}[c]{0.2\textwidth}
    \caption{Qualitative comparison of face generation results with other recent image generation approaches.}
	\label{fig:comparison}
  \end{minipage}
  \;
  \begin{minipage}[c]{0.72\textwidth}
    \includegraphics[width=1\linewidth]{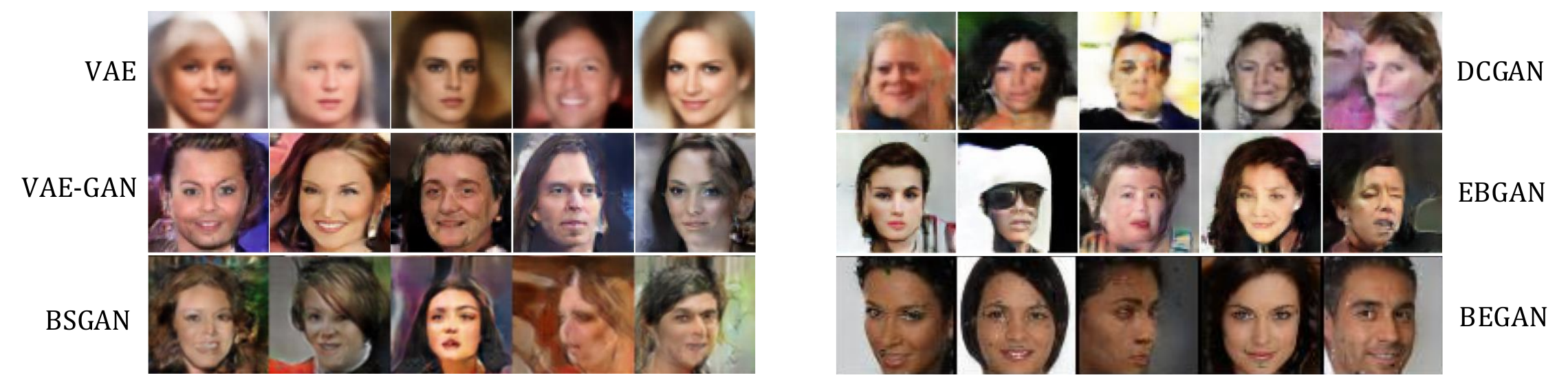}
  \end{minipage}\hfill
\end{figure*}

\subsection{Results}
Our sample generated face images are shown in Figures \ref{fig:results_64}, \ref{fig:results_128} and \ref{fig:reconstructions}. Our approach was able to generate a diverse set of face images, belonging to different age groups and containing a variety of attributes such as the blonde hair, smiling face and heavy make-up. The generated images contain both male and female examples, although we noticed a bias towards the female category due to a heavy representation of female celebrities in the celebA dataset. The generated samples also contain pose variations and a diverse set of facial expressions. 

The proposed method allowed us to easily scale up the resolution of generated samples while maintaining their visual quality (see Figure \ref{fig:results_128}). However, we noticed some smoothing effects when the model was trained to generate $128\times 128$ images compared to $64\times 64$ images.  Since the discriminator in our model is an auto-encoder, it is also of interest to visualize the reconstruction performance for both the real and fake images (Figure \ref{fig:reconstructions}). Such a reconstruction is not possible with the original GAN based model. We noticed that the generated samples are less complex, and are therefore reconstructed very well by the discriminator. In contrast, the real images are more challenging to reproduce and therefore suffer from over-smoothing by the generator. It also shows that the equilibrium between generator and discriminator is maintained till the end of the training process.

\subsection{Quantitative Comparisons}
For quantitative evaluation of our approach, we compute the Inception score proposed in \cite{salimans2016improved} for the CIFAR-10 dataset. The training procedure is carried out in an unsupervised manner for a fair comparison with the previous techniques listed in Table~\ref{tab:quan_comp}. Precisely, for every generated image, the Inception model \cite{szegedy2016rethinking} is used to  calculate $p(y|\mathbf{x})$, where $y$ denotes the class label. Good quality samples are expected to have $p(y|\mathbf{x})$ with low entropy (i.e., consistent predictions) and the marginalized distribution over all samples $p(y)$ with high entropy (i.e. diverse predictions). The inception score is defined as the combination of these two criterion: $\exp(\mathbb{E}[\mathrm{KL}(p(y|\mathbf{x})\parallel p(y))])$. This measure was found to match well with the human evaluations of the quality of generated samples. Our quantitative results show the significance of using a learned similarity metric which provides significant improvement over the BEGAN \cite{berthelot2017began} approach. The proposed model with the inclusion of VAE with additional constraints on the latent representations lead to the significantly better performance compared to the closely related BEGAN. For comparisons, we also report the oracle case with real images, and the case where semi-supervised learning is performed as proposed in \cite{salimans2016improved}. Our unsupervised model scores fairly close to the semi-supervised version of \cite{salimans2016improved}.

\subsection{Qualitative Comparisons}
We qualitatively compare our results with: \textbf{(i)} a VAE model comprising of an encoder and  a decoder \cite{kingma2013auto}, \textbf{(ii)} a deep convolutional GAN \cite{radford2015unsupervised}, \textbf{(iii)} combined VAE+GAN model \cite{larsen2015autoencoding}, \textbf{(iv)} energy based GAN \cite{zhao2016energy}, \textbf{(v)}  boundary equilibrium GAN \cite{berthelot2017began} and \textbf{(vi)} the boundary seeking GAN \cite{hjelm2017boundary}. Note that the results reported in \cite{berthelot2017began} are obtained by training on a much larger (but publicly unavailable) dataset comprising of 360k celebrity face images. For a fair comparison, we train their model on the CelebA dataset with the same parameter settings as reported by the authors. All other approaches were already trained on the CelebA dataset, therefore we take their reported results for comparison. Since the results for larger image sizes are not available for most of these approaches, we compare for the case of $64\times 64$ images (see Figure \ref{fig:comparison}).

The VAE model implemented using the convolutional encoder and decoder is trained using the pixel wise loss and therefore generates blurry images with high bias towards the frontalized head poses. The deep convolutional GAN and energy based GAN generate much sharp images but contain both local and global visual artifacts. The combination of VAE and GAN gives better results than the original VAE based model, however it still does not solve the blur and noise in the output images. The boundary equilibrium GAN generates fairly good quality images with a diverse set of facial attributes, however their model have problems dealing with side poses. Lastly, the boundary seeking GANs have high contrast close to face boundaries but the overall faces look unrealistic.

\begin{figure}[t]
	\centering
	\includegraphics[width=1\columnwidth]{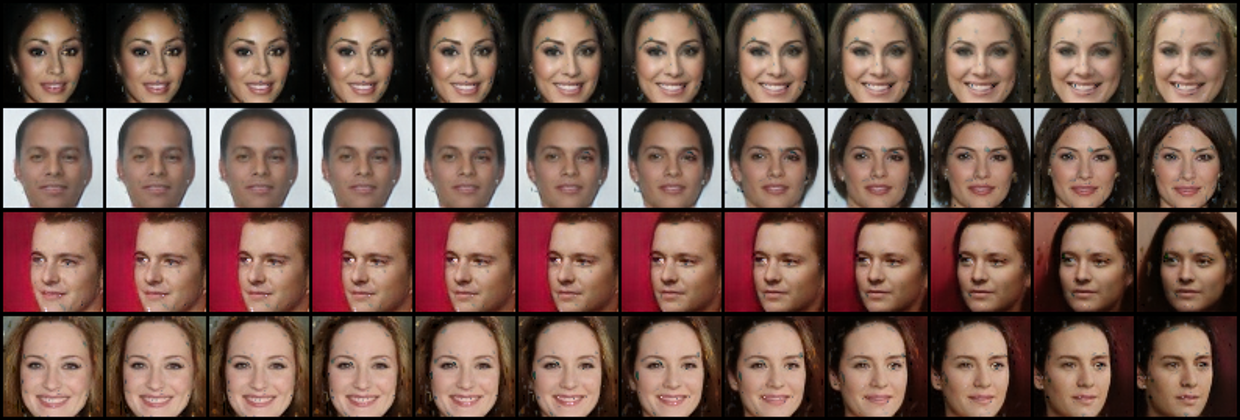}
	\caption{Interpolation between the two generated images ($64\times 64$) by moving in the latent space.}
	\label{fig:interpolation}
\end{figure}

\begin{figure}[t]
	\centering
	\includegraphics[width=1\linewidth,keepaspectratio=true]{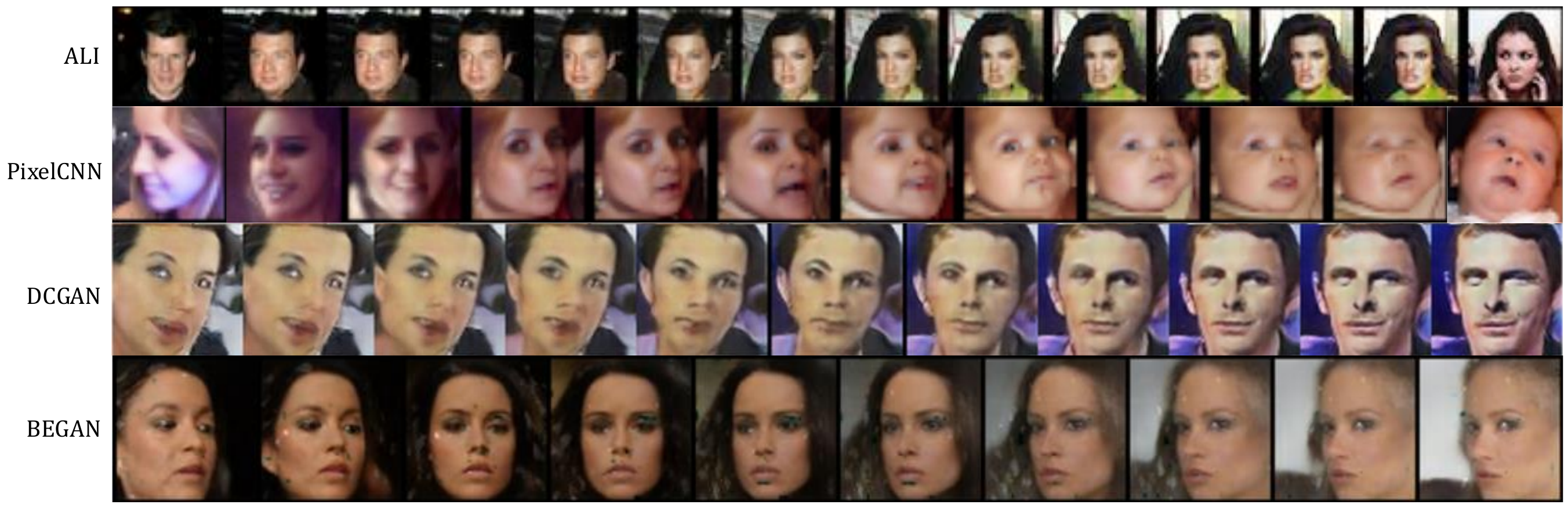}
	\caption{Comparison for continuity in latent space.}
	\label{fig:comparison_inter}
\end{figure}

\subsection{Exploring the Latent Space}\label{Exploring the Latent Space}
In order to validate that the learned generator has not merely memorized the training examples, we experiment the continuity in the latent space to check if the intermediate latent representations also correspond to realistic images. To this end, we find the latent vectors $z$ corresponding to two real images. We then interpolate between the two $z$ embeddings and show the corresponding images generated by the model. The results are reported in Figure \ref{fig:interpolation}. We note that there exists smooth transition between the real faces, even in cases where the two images are remarkably different.  This proves that the model has good generalization capability and has not memorized the image contents. Here, we also qualitatively compare with other image generation methods including ALI \cite{dumoulin2016adversarially}, PixelCNN \cite{van2016conditional}, DCGAN \cite{radford2015unsupervised} and BEGAN \cite{berthelot2017began} in Figure \ref{fig:comparison_inter}.

\begin{figure}[t] 
	\centering
	\includegraphics[width=0.85\columnwidth]{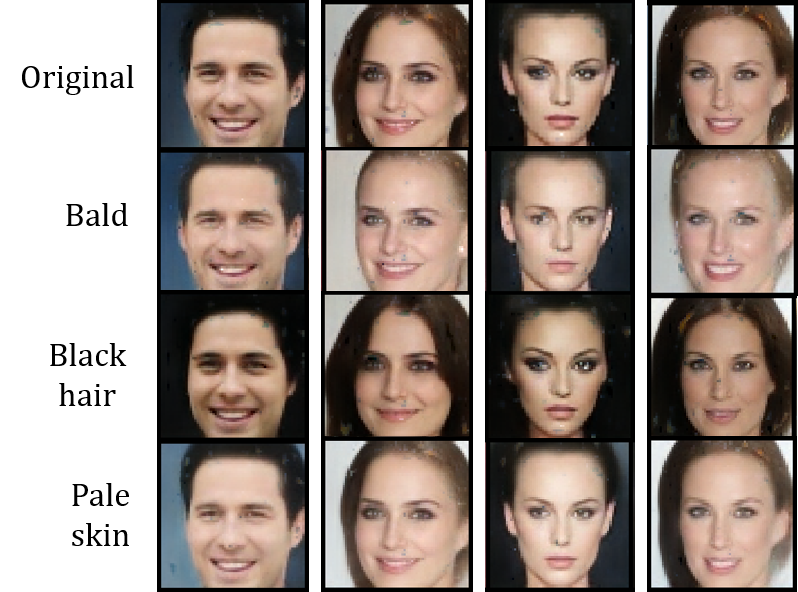}
	\caption{Latent space arithmetic results for example attributes. }
	\label{fig:latent_arith}
\end{figure}

A distinguishing feature of our model is the flexibility to control the generated images by appropriately modifying inputs in the latent space.  This is in contrast to regular GAN based models, which do not allow such a control over the latent space since the samples are randomly drawn from a uniform distribution. In this experiment, we aim to study the relationships between the latent vectors and the face attributes. For this purpose, we encode each image using the VAE and average the latent vectors to obtain a representation for each of the 40 semantically meaningful attributes in the celebA dataset. Similar to \cite{mikolov2013efficient}, we show that applying simple arithmetic operations in the latent space using the average response corresponding to each attribute results in the modified images with the desired attributes. Specifically, we calculate the average latent encoded representations of all images with and without a specific attribute respectively. Then a weighted version of the difference vector can be added to the newly generated image latent representation to add or remove a specific attribute. We show some examples of latent space arithmetic in Figure \ref{fig:latent_arith} where attributes such as bald, black hair and pale skin are clearly reflected on to the generated images.

\section{Conclusion}
This paper proposed a new approach to train VAEs by matching the data as well as the loss distributions of the real and fake images. This was achieved by a pair of auto-encoders which served as the generator and the discriminator in the adversarial training. Our model automatically learned a similarity metric defined in terms of latent representation obtained using the discriminator to enable generation of high-quality image outputs. This helped in overcoming the artifacts caused when only the data or loss distribution was matched between samples. Our method utilized a simple model architecture, was stable during training and easily scaled up to generate high dimensional perceptually realistic outputs. In future, we will explore the conditional image generation models to obtain even higher image resolutions while maintaining the photo-realistic quality. 

\section*{Acknowledgments} Thanks to NVIDIA Corporation for donation of the Titan X Pascal GPU used for this research.

{\small
\bibliographystyle{ieee}
\bibliography{egbib}
}

\end{document}